\pdfoutput=1
\documentclass[11pt]{article}

\usepackage[]{EMNLP2023}

\usepackage{times}
\usepackage{latexsym}

\usepackage[T1]{fontenc}
\usepackage[utf8]{inputenc}

\usepackage{microtype}

\usepackage{inconsolata}

\usepackage{url}
\usepackage{hyperref}
\usepackage{algorithm}
\usepackage{algorithmic}
\usepackage{listings}
\usepackage{newfloat}
\usepackage{listings}
\usepackage{titlesec}
\usepackage{verbatim}
\usepackage{multirow}
\usepackage{graphicx}
\usepackage{booktabs}
\usepackage{amssymb}
\usepackage{amsmath}
\usepackage{bm}
\usepackage{color, soul}
\usepackage{fontawesome5}
\usepackage{enumitem}
\usepackage[textsize=tiny]{todonotes}

\usepackage[capitalise]{cleveref}

\usepackage{tikz}
\usepackage[edges]{forest}
\definecolor{hidden-draw}{RGB}{205, 44, 36}
\definecolor{hidden-blue}{RGB}{194,232,247}
\definecolor{hidden-orange}{RGB}{243,202,120}
\definecolor{hidden-yellow}{RGB}{242,244,193}
\definecolor{tree-level-1}{RGB}{245,20,85}
\definecolor{tree-level-2}{RGB}{246,86,118}
\definecolor{tree-level-3}{RGB}{248,177,193}
\definecolor{tree-leaf}{RGB}{176,230,198}

\definecolor{Self}{RGB}{255,0,128}
\definecolor{Ensemble}{RGB}{0,127,255}
\definecolor{Iterative}{RGB}{153,51,255}

\definecolor{exemplar1}{RGB}{136,98,148}
\definecolor{exemplar2}{RGB}{148,210,242}
\definecolor{knowledge1}{RGB}{249,219,152}
\definecolor{knowledge2}{RGB}{255,245,220}

\usepackage{subcaption}
\usepackage{pgfplots}
\pgfplotsset{compat=1.17}
\usetikzlibrary{pgfplots.groupplots}

\usepackage[algo2e]{algorithm2e} 
\definecolor{myyellow}{RGB}{255, 204, 0}

\definecolor{wingreen}{rgb}{0,0.45,0.24}
\definecolor{losered}{rgb}{1.0,0.1,0.24}

\definecolor{lightcoral}{rgb}{0.94, 0.5, 0.5}
\definecolor{lightgreen}{rgb}{0.56, 0.93, 0.56}
\definecolor{harvestgold}{rgb}{0.85, 0.57, 0.0}
\definecolor{brightlavender}{rgb}{0.75, 0.58, 0.89}
\definecolor{capri}{rgb}{0.0, 0.75, 1.0}
\definecolor{carminepink}{rgb}{0.92, 0.3, 0.26}
\definecolor{celadon}{rgb}{0.67, 0.88, 0.69}
\definecolor{darkpastelgreen}{rgb}{0.01, 0.75, 0.24}

\newcommand{\methodname}[1]{Survey}
\newcommand{\basea}[1]{base}
\newcommand{\baseb}[1]{base}
\newcommand{\data}[1]{data}

\title{A Survey on Detection of LLMs-Generated Content}

\author{Xianjun Yang$^{1}$ \qquad Liangming Pan$^{1}$ \qquad Xuandong Zhao$^{1}$ \qquad Haifeng Chen$^{2}$ \\ \bf
Linda Petzold$^{1}$ \qquad William Yang Wang$^{1}$ \qquad Wei Cheng$^{2}$ \\
\texttt{\{xianjunyang, liangmingpan, xuandongzhao, petzold, wangwilliamyang\}@ucsb.edu}\\ \texttt{\{weicheng, haifeng\}@nec-labs.com}\\
$^{1}$ University of California, Santa Barbara \qquad $^{2}$ NEC Laboratories America, Princeton}

\begin{document}
\maketitle

\begin{abstract}
The burgeoning capabilities of advanced large language models (LLMs) such as ChatGPT have led to an increase in synthetic content generation with implications across a variety of sectors, including media, cybersecurity, public discourse, and education. As such, the ability to detect LLMs-generated content has become of paramount importance.
We aim to provide a detailed overview of existing detection strategies and benchmarks, scrutinizing their differences and identifying key challenges and prospects in the field, advocating for more adaptable and robust models to enhance detection accuracy. We also posit the necessity for a multi-faceted approach to defend against various attacks to counter the rapidly advancing capabilities of LLMs.
To the best of our knowledge, this work is the first comprehensive survey on the detection in the era of LLMs. We hope it will provide a broad understanding of the current landscape of LLMs-generated content detection, offering a guiding reference for researchers and practitioners striving to uphold the integrity of digital information in an era increasingly dominated by synthetic content. The relevant papers are summarized and will be consistently updated at \url{https://github.com/Xianjun-Yang/Awesome_papers_on_LLMs_detection.git}.
\end{abstract}

\section{Introduction}
With the rapid development of powerful AI tools, the risk of LLMs-generated content has raised considerable concerns, such as misinformation spread \citep{bian2023drop, hanley2023machine, pan2023risk}, fake news \citep{oshikawa2018survey, zellers2019defending, dugan2022real}, gender bias \citep{sun-etal-2019-mitigating}, education \citep{perkins2023game, vasilatos2023howkgpt}, and social harm \citep{kumar-etal-2023-language, yang2023shadow}. 

\begin{figure}\vspace{-0.2cm}
\centering
    \includegraphics[width=.49\textwidth]{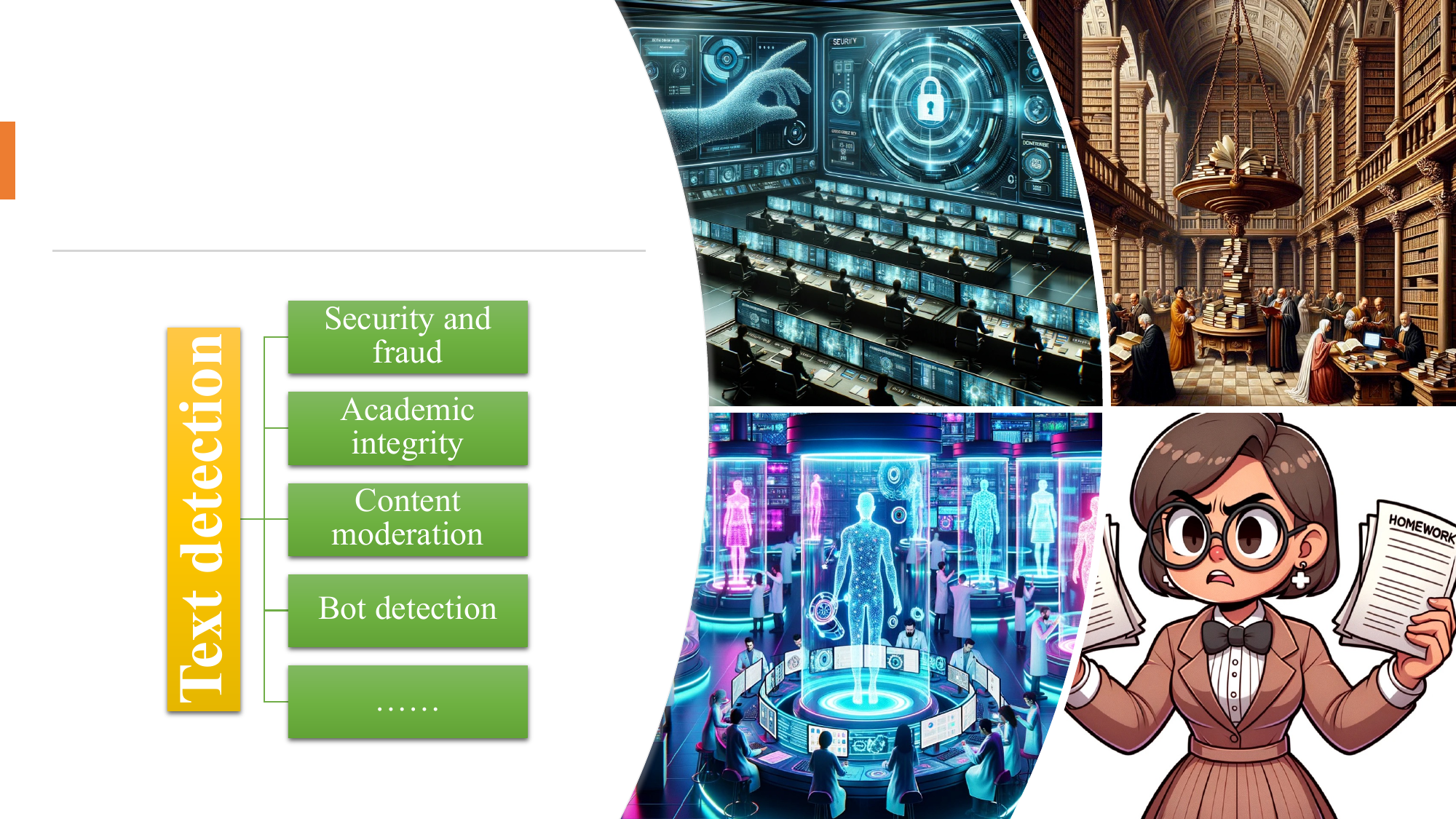}
    \caption{We list the four most representative scenarios where the detection of LLMs-generated content is of paramount importance. 
    }\label{fig:trend}
\end{figure}

In Figure \ref{fig:trend}, we show some topics regarding the threats brought by AI-written text on education, social media, elections, etc. We also find on the Google search trend, that the concerns about AI-written text have witnessed a significant increase since the release of the latest powerful Large Langue Models (LLMs) such as ChatGPT \citep{schulman2022chatgpt} and GPT-4 \citep{openai2023gpt4}. Humans are already unable to directly distinguish between LLMs- and human-written text, with the fast advancement of the model size, data scale, and AI-human alignment \citep{brown2020language, ouyang2022training}. Since then, the field of LLMs has made remarkable advancements, leading to substantial improvements in the generation of content across various NLP tasks \citep{qin2023chatgpt, yang2023exploring}. Alongside these advancements, there has been a proliferation of detection algorithms aimed at identifying LLMs-generated content. However, there remains a dearth of comprehensive surveys encompassing the latest methodologies, benchmarks, and attacks on LLMs-based detection systems.
\vspace{-0.1cm}

Actually, AI has revolutionized various modalities, encompassing image generation models like DALLE \citep{ramesh2021zero} and Imagen \citep{saharia2022photorealistic}, text generation models like ChatGPT and Bard, audio processing models such as MMS \citep{pratap2023scaling}, video generation models \citep{singer2022make}, and code generation models \citep{chen2021evaluating}.
In the past, there have been efforts to develop watermarking techniques for images \citep{wen2023tree, zhao2023recipe} and successful attacks against such techniques \citep{jiang2023evading}. Additionally, research has been conducted on the detection of online ChatBots  \citep{ chew2003baffletext, wang2023bot}. Specifically, this survey concentrates on the detection of text and code generated by LLMs, as well as the attacks.

The current most powerful commercial LLMs, e.g. Anthropic's Claude, OpenAI's ChatGPT, and GPT-4, usually adopt the decoder-only transformer \citep{vaswani2017attention} architecture. Those models have tens to hundreds of billions of parameters, are trained on a large collection of text, and are further tuned to align with human preferences. During inference, the text generation process involves using top-$k$ sampling \citep{fan-etal-2018-hierarchical}, nucleus sampling \citep{holtzmancurious} in conjunction with beam search. 
Concurrently, growing interests are shown to detectors, like the commercial tool GPTZero \citep{GPTZero}, or OpenAI's own detector \citep{AITextClassifier} since humans can be easily fooled by improvements in decoding methods \citep{ippolito2019automatic}. However, the misuse of detectors also raises protests from students on the unfair judgment on their homework and essays \citep{herbold2023ai, liu2023argugpt} and popular detectors perform poorly on code detection \citep{wang2023evaluating}.

Earlier work on text detection dates back to feature engineering \citep{badaskar2008identifying}. For example, GTLR \citep{gehrmann2019gltr} assumes the generated word comes from the top distribution on small LMs like BERT \citep{devlin-etal-2019-bert} or GPT-2 \citep{radford2019language}. Recently, there has been an increasing focus on detecting ChatGPT \citep{weng2023towards, liu2023argugpt, desaire2023chatgpt}, to mitigate ChatGPT misuse or abuse \citep{sison2023chatgpt}. In particular, it has recently been called for regulation\footnote{https://www.nytimes.com/2023/05/16/technology/openai-altman-artificial-intelligence-regulation.html} on powerful AI like ChatGPT usage \citep{hacker2023regulating, wahle2023ai}.

Therefore, we firmly believe that the timing is ideal for a comprehensive survey on the detection of LLMs-generated content. It would serve to invite further exploration of detection approaches, offer valuable insights into the strengths and weaknesses of previous research, and highlight potential challenges and opportunities for the research community to address.
Our paper is organized as follows: we first briefly describe the problem formulation, including the task definition, metrics, and datasets in Section \ref{sec: formulation}. In Section \ref{sec: Scenarios}, we classify detection by their working mechanism and scope of application. 
In section \ref{sec:Methodologies}, we summarize the three popular detection methods: training-based, zero-shot and watermarking. 
We also investigate various attacks in Section \ref{sec: attack} since defending against attacks is of increasing importance and point out some challenges in Section \ref{sec: challenge}. 
Finally, in Section \ref{sec: outlook} we provide additional insights into this topic on potential future directions, as well as the conclusion in Section \ref{sec: conclusion}.

\vspace{-0.15cm}
\section{Problem formulation}\label{sec: formulation}
\vspace{-0.15cm}
\subsection{Overview}
We refer to any textual outputs from LLMs following specific inputs as LLMs-Generated Content. It can be generally classified into natural languages like news, essays, reviews, and reports, or programming languages like codes of Python, C++, and Java. Current research usually aims at the detection of content with moderate length and specific topics. It is meaningless to detect a short sentence describing some facts like \textit{EMNLP started in 1996} or simple coding question \textit{ def hello\_world(): print('Hello World')}, to be human or AI written. 

Formally, consider an LLM denoted as $LLM$, which generates a candidate text $S$ of length $|S|$ based on an input prompt. Let $f()$ represent a potential detector we aim to use for classification, assigning $f(S)$ to $0$ or $1$, where $0$ and $1$ signify human or machine, respectively. The $LLM$ can be classified into unknown (Black-box), fully known (White-box), or partially known (known model name with unknown model parameters) to the detectors.
In practice, we are usually given a candidate corpus $C$ comprising both human and LLMs-generated content to test $f()$. 

Apart from the standard definition, machine-generated content can undergo additional modifications in practical scenarios, including rephrasing by humans or other AI models. Besides, it is also possible that the candidate text is a mix of human and machine-written text. For example, the first several sentences are written by humans, and the remaining parts by machines, or vice versa. When a text undergoes revisions, the community often perceives it as paraphrasing and treats it as either machine- or human-generated text, depending on the extent of these modifications and the intent behind them. 
However, it is important to highlight that if a substantial majority of the text is authored by humans, or if humans have extensively revised machine-generated text, it becomes challenging to maintain the assertion that the text is purely machine-generated. Hence, in this survey, we adhere to the traditional definition by considering machine-generated content as text that has not undergone significant modifications, and we consistently classify such text as machine-generated.

\subsection{Metrics} 
Previous studies \citep{mitchell2023detectgpt, sadasivan2023can} predominantly used the Area Under the Receiver Operating Characteristic (AUROC) score to gauge the effectiveness of detection algorithms. 
As a binary classification problem, AUROC shows the results under different thresholds, and the F1 score is also helpful. 
\citet{krishna2023paraphrasing, yang2023dna} suggest that AUROC may not consistently provide a precise evaluation, particularly as the AUROC score nears the optimal limit of 1.0 since two detectors with identical AUROC score of $0.99$ could exhibit substantial variations in detection quality from a user's perspective. 
From a practical point of view, ensuring a high True Positive Rate (TPR) is imperative while keeping the False Positive Rate (FPR) to a minimum. As such, current research \citep{krishna2023paraphrasing, yang2023dna} both report TPR scores at a fixed 1\% FPR, along with the AUROC. Other work \citep{sadasivan2023can} also refer to Type I and Type II errors following the binary hypothesis test and even report TPR at $10^{-6}$ FPR \citep{fernandez2023three}.

\subsection{Datasets} \label{02: datasets}%\vspace{-0.1cm}

In this section, we discuss the common datasets used for this task. The corpus is usually adopted from previous NLP tasks, and reconstructed by prompting LLMs to generate new outputs as candidate machine-generated text. Usually, there are two prompting methods: 1). prompting LLMs with the questions in some question-answering datasets.
2). prompting LLMs with the first 20 to 30 tokens to continue writing in datasets without specific questions. Specifically, several datasets have been compiled and utilized in the field. Some noteworthy datasets include TURINGBENCH \citep{uchendu-etal-2021-turingbench-benchmark}, HC3 \citep{guo2023close}, CHEAT \citep{yu2023cheat}, Ghostbuster \citep{verma2023ghostbuster}, OpenGPTText \citep{chen2023gpt}, M4 \citep{wang2023m4}, MGTBench \citep{he2023mgtbench}, and MULTITuDE \citep{macko2023multitude}
and some other datasets not explicitly built for detection have also been used, such as C4 \citep{2019t5}, shareGPT \footnote{https://sharegpt.com/}, and alpaca \citep{alpaca}, as summarized in Table \ref{table:datasets}. For text detection, we only list datasets explicitly built for detection, while some general datasets like C4 \citep{2019t5} or alpaca \citep{alpaca} can also be used. For code detection, we only list datasets that have been used in previous code detection work \citep{lee2023wrote, yang2023zero}. And other code\-generation corpora can also be adopted. The detailed description is included in Appendix \ref{sec:datasets}.
\begin{table*}[t]
\centering
\resizebox{0.99\textwidth}{!}
{
\begin{tabular}{ccccc}
\hline
\textbf{Datasets} & \textbf{Length} & \textbf{Size}  & \textbf{Data type} & \textbf{\#Language} \\ \hline
TuringBench   \citeyearpar{uchendu-etal-2021-turingbench-benchmark} & 100$\sim$400 & 200K    & News articles & 1  \\ 
\hline
HC3 \citeyearpar{guo2023close} & 100$\sim$250  & 44,425  & Reddit, Wikipedia, medicine and finance  & 2 \\ 
\hline
CHEAT \citeyearpar{yu2023cheat}  & 100$\sim$300 & 35,304 & Academical abstracts & 1 \\ \hline
Ghostbuster   \citeyearpar{verma2023ghostbuster} & 200$\sim$1200  &  12,685  & Student essays, creative fiction, and news & 1 \\ 
\hline
GPT-Sentinel \citeyearpar{chen2023gpt} & 100$\sim$400 & 29,395 & OpenWebText \citeyearpar{gokaslan2023openwebtext}  & 1\\ \hline
M4 \citeyearpar{wang2023m4} & 200-300  & 122,481& Multi-domains & 6 \\ 
\hline
MGTBench \citeyearpar{he2023mgtbench} & 10$\sim$200 & 2,817 & Question-answering datasets & 1 \\ 
\hline
HC3 Plus \citeyearpar{su2023hc3} & 100$\sim$250 & 214,498 & Summarization, translation, and paraphrasing  & 2 \\ 
\hline
MULTITuDE \citeyearpar{macko2023multitude} & 150$\sim$400 & 74,081 & MassiveSumm \citeyearpar{varab-schluter-2021-massivesumm}  & 11 \\ 
\hline
HumanEval \citeyearpar{chen2021evaluating} & $\sim$181 & 164 & Code Exercise  & 1 \\ 
\hline
APPS \citeyearpar{hendrycks2021measuring} & $\sim$474 & 5,000 & Code Competitions  & 1 \\ 
\hline
CodeContests \citeyearpar{li2022competition} & $\sim$2239 & 165 & Code Competitions  & 6 \\ 
\hline
\end{tabular}
}\caption{A summarization of the detection datasets. Length is reported in the number of words for text and characters for codes. \#Language represents the number of types of natural languages for text and programming languages for codes. }\label{table:datasets}\vspace{-0.2cm}
\end{table*}
% MGTBench: TruthfulQA, SQuAD, NarrativeQA

\textbf{Data Contamination.}
Despite those released standard datasets, we argue that static evaluation benchmarks might not be desirable for this problem with the rapid progress of LLMs trained, tuned, or aligned on large amounts of data across the whole internet. 
On the one hand, \citet{aaronson2022my} mentioned that some text from Shakespeare or the Bible is often classified as AI-generated because such classic text is frequently used in the training datasets for generative language models.
On the other hand, many detectors did not fully disclose their training data, especially commercial tools like GPTZero \citep{GPTZero}. It is natural to worry that those standard evaluation benchmarks would face a serious test data contamination problem, considering the commercial detectors would consistently improve their products for profits.
So, with the rapid evolution of LLMs and detectors, the traditional paradigm of providing standard benchmarks might no longer be suitable for AI-generated text detection. We provide a unique solution to this:
% faSmile  faStar faLaughSquint faCrown

\textcolor{myyellow}{\faCrown} \textbf{Utilize the most latest human-written content to reduce data contamination problem by collecting such content from the most updated open-source websites, which themselves explicitly forbid posting AI-written posts. }

%\vspace{-0.3cm}
\section{Detection Scenarios}\label{sec: Scenarios}

\begin{figure}
\centering
\resizebox{0.48\textwidth}{!}{%
    \includegraphics{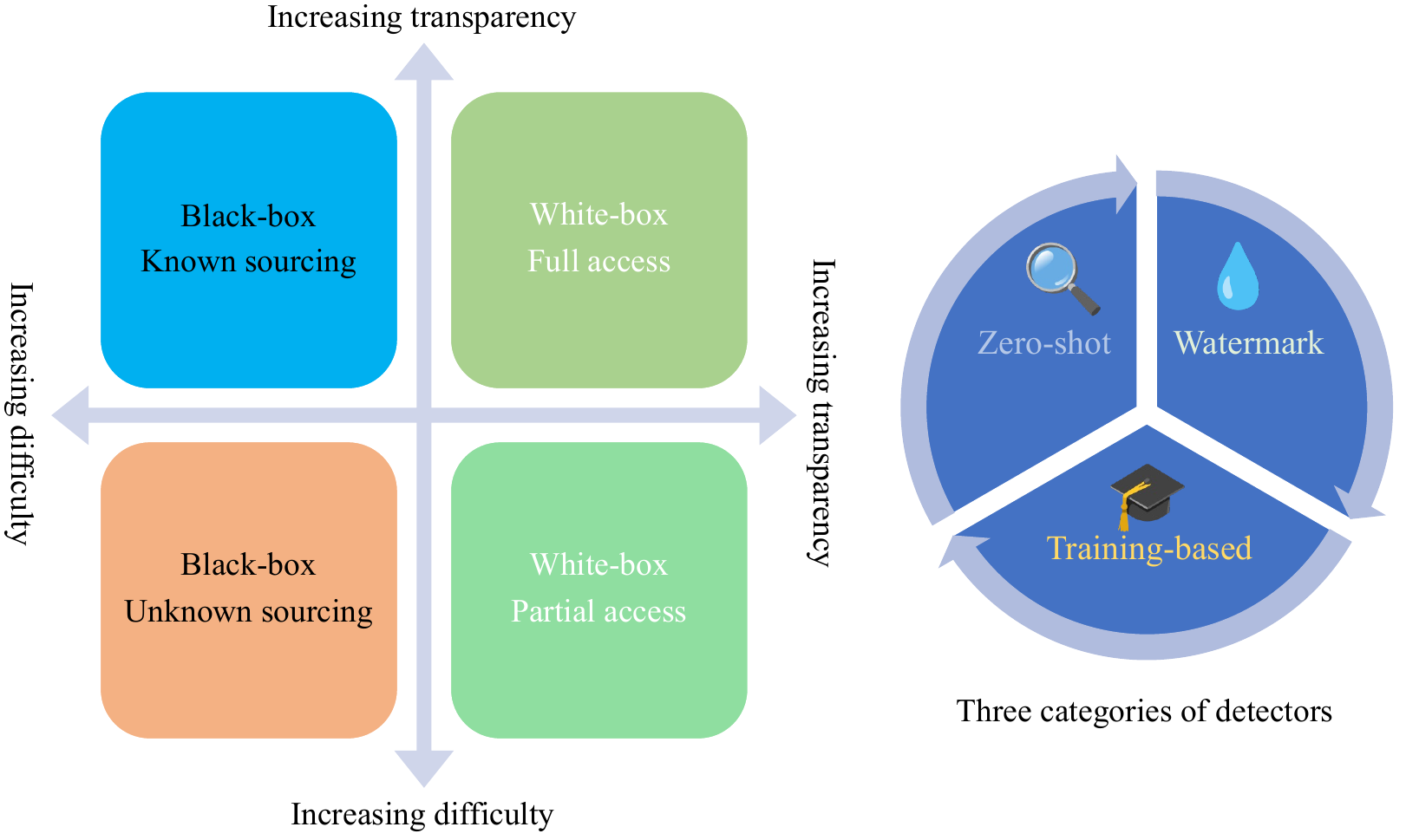}%
}
\caption{Three categories of detectors and four detection scenarios: as the transparency decreases, the detection difficulty increases. }
\label{fig: classification}\vspace{-0.4cm}
\end{figure}\vspace{-0.3cm}

The findings of previous research, such as \citep{gehrmann-etal-2019-gltr} and \citep{dugan2022real}, highlight the general difficulty faced by humans in distinguishing between human- and machine-generated text, motivating the development of automatic solutions. The detection process can be classified into black-box or white-box detection based on whether the detector has access to the source model output logits. In black-box detection, there are two distinct cases:
1). when the source model name is known, such as \texttt{GPT-4}; 2). when the source model name is unknown, and the content might have been generated by models like \texttt{GPT-4}, \texttt{Bard}, or other undisclosed models.
On the other hand, white-box detection also encompasses two cases: 1). the detector only has access to the model's output logits or partial logits, such as the top-5 token log probability in \texttt{text-davinci-003}; 2). the detector has access to the entire model weights.
Table \ref{fig: classification} shows four categories according to application scenarios and three detector methods. 
Specifically, we can categorize the usage of detecting LLM-generated content into four distinct scenarios based on their application:
These categorizations highlight the different levels of information available to the detectors, ranging from limited knowledge to complete access and demonstrate the various scenarios encountered in detecting machine-generated content.

\subsection{Black-Box Detection with Unknown Model Source}
This scenario closely resembles real-world applications, particularly when users, such as students, utilize off-the-shelf AI services to assist them in writing their essays. In such cases, teachers are often unaware of the specific AI service being employed. Consequently, this situation poses the greatest challenge as very limited information is available to identify instances of deception.

\subsection{Black-Box Detection with Known Model Source}
In this scenario, we possess knowledge regarding the specific model from which the text originates, yet we lack access to its underlying parameters. This aspect carries considerable significance due to the market domination of major language model providers such as OpenAI and Google. Many users rely heavily on their services, enabling us to make informed assumptions about the model sources.

\subsection{White-Box Detection with Full Model Parameters}
While access to the most powerful LLMs, such as Anthropic's Claude or OpenAI's ChatGPT, is typically limited, assuming full access to the model parameters is an active research area. This approach is reasonable, considering that researchers often encounter resource constraints, making it challenging to experiment with large-scale models. 
For instance, watermarking-based methods \citep{kirchenbauer2023watermark} typically require full access to the model parameters. This technique manipulates the next token prediction at each sampling position by modifying the distribution. Although this approach necessitates access to the complete model parameters, it has shown promise and could potentially be adapted for practical use.

\subsection{White-Box Detection with Partial Model Information}
This corresponds to the scenarios when only the partial model outputs, like the top-5 token logits are provided by \texttt{text-davinci-003}. Previous work like DetectGPT \citep{mitchell2023detectgpt} and DNA-GPT \citep{yang2023dna} both utilize such probability to perform detection.

\subsection{Model Sourcing}
Furthermore, another aspect related to detection goes beyond distinguishing between human and machine-generated content. This task involves determining which specific model may have generated the content and is referred to as authorship attribution \citep{uchendu-etal-2020-authorship}, origin tracing \citep{li2023origin}, or model sourcing \citep{yang2023dna}. We consider this task as a special scenario since it is slightly different from other detection tasks.

\begin{figure*}[tp]
    \centering
    \tikzstyle{my-box}=[
    rectangle,
    draw=hidden-draw,
    rounded corners,
    text opacity=1,
    minimum height=1.5em,
    minimum width=5em,
    inner sep=2pt,
    align=center,
    fill opacity=.5,
    ]
    \tikzstyle{leaf}=[my-box, minimum height=1.5em,
        fill=hidden-orange!60, text=black, align=left,font=\scriptsize,
        inner xsep=2pt,
        inner ysep=4pt,
    ]
    \resizebox{0.99\textwidth}{!}{
        \begin{forest}
            forked edges,
            for tree={
                grow=east,
                reversed=true,
                anchor=base west,
                parent anchor=east,
                child anchor=west,
                base=left,
                font=\small,
                rectangle,
                draw=hidden-draw,
                rounded corners,
                align=left,
                minimum width=4em,
                edge+={darkgray, line width=1pt},
                s sep=3pt,
                inner xsep=2pt,
                inner ysep=3pt,
                ver/.style={rotate=90, child anchor=north, parent anchor=south, anchor=center},
            },
            where level=1{text width=3.9em,font=\scriptsize,}{},
            where level=2{text width=4.9em,font=\scriptsize,}{},
            where level=3{text width=4.6em,font=\scriptsize,}{},
            where level=4{text width=6.8em,font=\scriptsize,}{},
            where level=5{text width=10.8em,font=\scriptsize,}{},
            [
                LLMs-generated content detection, draw=gray, color=gray!100, fill=gray!15, very thick, text=black, ver
                [
                   Training-based \\ Methods \\ (\S \ref{sec:training-method}), color=lightcoral!100, fill=lightcoral!15, very thick, text=black
                    [
                        Black-box (\S \ref{sec:training-black-box}), color=lightcoral!100, fill=lightcoral!15, very thick, text=black
                        [
                            Known Source, color=lightcoral!100, fill=lightcoral!15, very thick, text=black
                            [
                                Mixed sources, color=lightcoral!100, fill=lightcoral!15, very thick, text=black
                                    [
                                        OpenAI text classifier \citep{AITextClassifier}{,}
                                        GPTZero \citep{GPTZero}{,}
                                        G$^3$Detector \citep{zhan2023g3detector}{,} \\
                                        GPT-Sentinel \citep{chen2023gpt}
                                        , leaf, text width=25em, color=lightcoral!100, fill=lightcoral!15, very thick, text=black
                                    ]
                            ]
                            [
                                Mixied decoding, color=lightcoral!100, fill=lightcoral!15, very thick, text=black
                                    [
                                        \citep{ippolito-etal-2020-automatic}{,}
                                        GPT-Pat \citep{yu2023gpt}
                                        , leaf, text width=25em, color=lightcoral!100, fill=lightcoral!15, very thick, text=black
                                    ]
                            ]
                            [
                                 Mixed strategies, color=lightcoral!100, fill=lightcoral!15, very thick, text=black
                                    [
                                        Graph structure and contrastive learning, color=lightcoral!100, fill=lightcoral!15, very thick, text=black
                                        [
                                            CoCo \citep{liu2022coco}{,}
                                            , leaf, text width=12.7em, color=lightcoral!100, fill=lightcoral!15, very thick, text=black
                                        ]
                                    ]    
                                    [
                                        Proxy perplexity, color=lightcoral!100, fill=lightcoral!15, very thick, text=black
                                        [
                                            LLMDet \citep{wu2023llmdet}
                                            , leaf, text width=12.7em, color=lightcoral!100, fill=lightcoral!15, very thick, text=black
                                        ]
                                    ]
                                    [
                                        Positive unlabeled training, color=lightcoral!100, fill=lightcoral!15, very thick, text=black
                                        [
                                            MPU \citep{tian2023multiscale}
                                            , leaf, text width=12.7em, color=lightcoral!100, fill=lightcoral!15, very thick, text=black
                                        ]
                                    ]
                                    [
                                        Adversarial training, color=lightcoral!100, fill=lightcoral!15, very thick, text=black
                                        [
                                            RADAR \citep{hu2023radar}
                                            , leaf, text width=12.7em, color=lightcoral!100, fill=lightcoral!15, very thick, text=black
                                        ]
                                    ]
                            ]
                        ]
                        [
                            Unknown Source, color=lightcoral!100, fill=lightcoral!15, very thick, text=black
                            [
                                Cross-domain transfer, color=lightcoral!100, fill=lightcoral!15, very thick, text=black
                                [
                                    \citep{pu2023zero}{,}
                                    GPTZero \citep{GPTZero}{,}
                                    Conda \citep{Bhattacharjee2023ConDACD}{,}\\
                                    Model family \citep{Antoun2023FromTT}
                                    , leaf, text width=25em, color=lightcoral!100, fill=lightcoral!15, very thick, text=black
                                ]
                            ]
                            [
                                Surrogate model, color=lightcoral!100, fill=lightcoral!15, very thick, text=black
                                [
                                    Ghostbuster \citep{verma2023ghostbuster}{,}
                                    , leaf, text width=25em, color=lightcoral!100, fill=lightcoral!15, very thick, text=black
                                ]
                            ]
                            [
                                Detection in the wild, color=lightcoral!100, fill=lightcoral!15, very thick, text=black
                                [
                                    Deepfake text detection \citep{li2023deepfake}{,}
                                    , leaf, text width=25em, color=lightcoral!100, fill=lightcoral!15, very thick, text=black
                                ]
                            ]
                        ]
                    ]
                    [
                        White-box (\S \ref{sec:training-white-box}), color=lightcoral!100, fill=lightcoral!15, very thick, text=black
                        [
                            Full access, color=lightcoral!100, fill=lightcoral!15, very thick, text=black
                            [   
                                Word rank , color=lightcoral!100, fill=lightcoral!15, very thick, text=black
                                [
                                    GLTR \citep{gehrmann2019gltr}{,}
                                    , leaf, text width=25em, color=lightcoral!100, fill=lightcoral!15, very thick, text=black
                                ]
                            ]
                        ]
                        [
                            Partial access, color=lightcoral!100, fill=lightcoral!15, very thick, text=black
                            [
                                Logits as waves, color=lightcoral!100, fill=lightcoral!15, very thick, text=black
                                [
                                    SeqXGPT \citep{wang2023seqxgpt}
                                    , leaf, text width=25em, color=lightcoral!100, fill=lightcoral!15, very thick, text=black
                                ]
                            ]
                            [
                                 Contrastive logits feature, color=lightcoral!100, fill=lightcoral!15, very thick, text=black
                                    [
                                        Sniffer \citep{li2023origin}
                                        , leaf, text width=25em, color=lightcoral!100, fill=lightcoral!15, very thick, text=black
                                    ]
                            ] 
                        ]
                    ]
                ]
                [
                    Zero-shot \\ Methods \\ (\S \ref{sec:Zero-Shot}), color=lightgreen!100, fill=lightgreen!15, very thick, text=black
                    [
                        Black-box (\S \ref{zero-shot: black}), color=lightgreen!100, fill=lightgreen!15, very thick, text=black
                        [
                            Known Source, color=lightgreen!100, fill=lightgreen!15, very thick, text=black
                            [
                                Database Retrieval, color=lightgreen!100, fill=lightgreen!15, very thick, text=black
                                [
                                    \citep{krishna2023paraphrasing}
                                    , leaf, text width=25em, color=lightgreen!100, fill=lightgreen!15, very thick, text=black
                                ]
                            ]
                            [
                                Uncommon n-grams, color=lightgreen!100, fill=lightgreen!15, very thick, text=black
                                [
                                     \citep{grechnikov2009detection}{,} \citep{badaskar2008identifying}
                                     , leaf, text width=25em, color=lightgreen!100, fill=lightgreen!15, very thick, text=black
                                ]
                            ]
                            [
                                  Probability curve , color=lightgreen!100, fill=lightgreen!15, very thick, text=black
                                [
                                    DetectGPT \citep{mitchell2023detectgpt}
                                    , leaf, text width=25em, color=lightgreen!100, fill=lightgreen!15, very thick, text=black
                                ]
                            ]
                            [
                                N-gram divergence, color=lightgreen!100, fill=lightgreen!15, very thick, text=black
                                [
                                    DNA-GPT \citep{yang2023dna}
                                    , leaf, text width=25em, color=lightgreen!100, fill=lightgreen!15, very thick, text=black
                                ]
                            ]
                            [
                                Smaller model as a proxy, color=lightgreen!100, fill=lightgreen!15, very thick, text=black
                                [
                                    \citep{mireshghallah2023smaller}
                                    , leaf, text width=25em, color=lightgreen!100, fill=lightgreen!15, very thick, text=black
                                ]
                            ]
                            [
                                Codes detection, color=lightgreen!100, fill=lightgreen!15, very thick, text=black
                                [
                                    DetectGPT4Code \citep{yang2023zero}
                                    , leaf, text width=25em, color=lightgreen!100, fill=lightgreen!15, very thick, text=black
                                ]
                            ]
                        ]
                        [
                            Unknown Source, color=lightgreen!100, fill=lightgreen!15, very thick, text=black
                            [
                                Intrinsic dimension, color=lightgreen!100, fill=lightgreen!15, very thick, text=black
                                [
                                    Persistent homology dimension estimator \citep{tulchinskii2023intrinsic}, color=lightgreen!100, fill=lightgreen!15, very thick, text=black
                                    , leaf, text width=25em, color=lightgreen!100, fill=lightgreen!15, very thick, text=black
                                ]
                            ]
                        ]
                    ]
                    [
                        White-box (\S \ref{zero-shot: white}), color=lightgreen!100, fill=lightgreen!15, very thick, text=black
                        [
                            Full access, color=lightgreen!100, fill=lightgreen!15, very thick, text=black
                            [   
                                Log-Rank ratio, color=lightgreen!100, fill=lightgreen!15, very thick, text=black
                                [
                                    DetectLLM-LRR \citep{su2023detectllm}
                                    , leaf, text width=25em, color=lightgreen!100, fill=lightgreen!15, very thick, text=black
                                ]
                            ]
                        ]
                        [
                            Partial access, color=lightgreen!100, fill=lightgreen!15, very thick, text=black
                            [
                                Traditional methods, color=lightgreen!100, fill=lightgreen!15, very thick, text=black
                                [
                                    Entropy, color=lightgreen!100, fill=lightgreen!15, very thick, text=black
                                    [
                                        \citep{lavergne2008detecting}
                                        , leaf, text width=12.7em, color=lightgreen!100, fill=lightgreen!15, very thick, text=black
                                    ]
                                ]
                                [
                                    Perplexity, color=lightgreen!100, fill=lightgreen!15, very thick, text=black
                                    [
                                        \citep{beresneva2016computer} 
                                        , leaf, text width=12.7em, color=lightgreen!100, fill=lightgreen!15, very thick, text=black
                                    ]
                                ]
                                [
                                    Log probability, color=lightgreen!100, fill=lightgreen!15, very thick, text=black
                                    [
                                        \citep{solaiman2019release}
                                        , leaf, text width=12.7em, color=lightgreen!100, fill=lightgreen!15, very thick, text=black
                                    ]
                                ]
                            ]
                            [
                                Recent methods, color=lightgreen!100, fill=lightgreen!15, very thick, text=black
                                [
                                    Probability curvature on perturbations, color=lightgreen!100, fill=lightgreen!15, very thick, text=black
                                    [
                                        DetectGPT \citep{mitchell2023detectgpt}
                                        , leaf, text width=12.7em, color=lightgreen!100, fill=lightgreen!15, very thick, text=black
                                    ]
                                ]
                                [
                                     Conditional probability divergence, color=lightgreen!100, fill=lightgreen!15, very thick, text=black
                                    [
                                        DNA-GPT \citep{yang2023dna}
                                        , leaf, text width=12.7em, color=lightgreen!100, fill=lightgreen!15, very thick, text=black
                                    ]
                                ]
                                [
                                     Conditional probability curvature, color=lightgreen!100, fill=lightgreen!15, very thick, text=black
                                    [
                                        Fast-DetectGPT \citep{bao2023fast}
                                        , leaf, text width=12.7em, color=lightgreen!100, fill=lightgreen!15, very thick, text=black
                                    ]
                                ]
                                [
                                     Uniform information density, color=lightgreen!100, fill=lightgreen!15, very thick, text=black
                                        [
                                            GPT-who \citep{venkatraman2023gptwho}
                                            , leaf, text width=12.7em, color=lightgreen!100, fill=lightgreen!15, very thick, text=black
                                        ]
                                ]
                                [
                                     Bayesian surrogate model, color=lightgreen!100, fill=lightgreen!15, very thick, text=black
                                        [
                                            \citep{deng2023efficient}
                                            , leaf, text width=12.7em, color=lightgreen!100, fill=lightgreen!15, very thick, text=black
                                        ]
                                ]
                            ]
                        ]
                    ]
                ]
                [
                    Watermarking \\ Methods \\ (\S \ref{sec:Watermarking}), color=cyan!100, fill=cyan!15, very thick, text=black
                    [
                        Black-box (\S \ref{watermark:black}), color=cyan!100, fill=cyan!15, very thick, text=black
                        [
                            Known Source, color=cyan!100, fill=cyan!15, very thick, text=black
                            [
                                Traditional methods, color=cyan!100, fill=cyan!15, very thick, text=black
                                [
                                    Paraphrasing \citep{atallah2003natural}{,} 
                                    Syntax tree manipulations \citep{topkara2005natural}{,}\\
                                    \citep{meral2009natural}{,}
                                    Synonym substitution \citep{topkara2006hiding}
                                    , leaf, text width=25em, color=cyan!100, fill=cyan!15, very thick, text=black
                                ]
                            ]
                            [
                                Latest methods, color=cyan!100, fill=cyan!15, very thick, text=black
                                [
                                    BERT-based lexical \citep{yang2022tracing}
                                    and synonyms \citep{yang2023watermarking} substitution 
                                    , leaf, text width=25em, color=cyan!100, fill=cyan!15, very thick, text=black
                                ]
                            ]
                        ]
                    ]
                    [
                        White-box (\S \ref{watermark:white}), color=cyan!100, fill=cyan!15, very thick, text=black
                        [
                            Known Source, color=cyan!100, fill=cyan!15, ultra thin, text=black
                            [
                                 Training-free watermark, color=cyan!100, fill=cyan!15, very thick, text=black
                                [
                                     Exponential minimum sampling , color=cyan!100, fill=cyan!15, very thick, text=black
                                    [
                                        \citep{aaronson2022my}
                                        , leaf, text width=12.7em, color=cyan!100, fill=cyan!15, very thick, text=black
                                    ]
                                ]
                                [
                                    Hashing of blocks  , color=cyan!100, fill=cyan!15, very thick, text=black
                                    [
                                        \citep{christ2023undetectable}
                                        , leaf, text width=12.7em, color=cyan!100, fill=cyan!15, very thick, text=black
                                    ]
                                ]
                                [
                                    Logits deviation w/ green-red list, color=cyan!100, fill=cyan!15, very thick, text=black
                                    [
                                        Soft watermark \citep{kirchenbauer2023watermark}
                                        , leaf, text width=12.7em, color=cyan!100, fill=cyan!15, very thick, text=black
                                    ]
                                ]
                                [
                                     Logits deviation w/ fixed split, color=cyan!100, fill=cyan!15, very thick, text=black
                                    [
                                        Unigram-Watermark \citep{Zhao2023ProvableRW}
                                        , leaf, text width=12.7em, color=cyan!100, fill=cyan!15, very thick, text=black
                                    ]
                                ]
                                [
                                     Sampling w/ randomized number, color=cyan!100, fill=cyan!15, very thick, text=black
                                    [
                                        \citep{Kuditipudi2023RobustDW}
                                        , leaf, text width=12.7em, color=cyan!100, fill=cyan!15, very thick, text=black
                                    ]
                                ]
                                [
                                     Sentence-level w/ rejection sampling, color=cyan!100, fill=cyan!15, very thick, text=black
                                    [
                                        SemStamp \citep{hou2023semstamp}{,} 
                                        , leaf, text width=12.7em, color=cyan!100, fill=cyan!15, very thick, text=black
                                    ] 
                                ] 
                                [
                                     Reweight strategy w/ ciphers, color=cyan!100, fill=cyan!15, very thick, text=black
                                    [
                                        DiPmark \citep{wu2023dipmark}{,} 
                                        , leaf, text width=12.7em, color=cyan!100, fill=cyan!15, very thick, text=black % , 
                                    ] 
                                ]
                            ]
                            [   
                                Training-based watermark, color=cyan!100, fill=cyan!15, very thick, text=black
                                [
                                     Logits deviation w/ semantic embeddings, color=cyan!100, fill=cyan!15, very thick, text=black
                                    [
                                        Training-free \citep{fu2023watermarking}{,} \\
                                        Training-based \citep{liu2023semantic}
                                        , leaf, text width=12.7em, color=cyan!100, fill=cyan!15, very thick, text=black
                                    ] 
                                ] 
                                [
                                     Message encoding w/ reparameterization, color=cyan!100, fill=cyan!15, very thick, text=black
                                    [
                                        REMARK-LLM \citep{zhang2023remark}{,} 
                                        , leaf, text width=12.7em, color=cyan!100, fill=cyan!15, very thick, text=black 
                                    ] 
                                ]
                            ]
                            [
                                Multi-bit watermark, color=cyan!100, fill=cyan!15, very thick, text=black
                                [
                                    Invariant Features, color=cyan!100, fill=cyan!15, very thick, text=black
                                    [
                                        \citep{yoo-etal-2023-robust}{,}
                                        , leaf, text width=12.7em, color=cyan!100, fill=cyan!15, very thick, text=black
                                    ]
                                ]
                                [
                                    Color-listing, color=cyan!100, fill=cyan!15, very thick, text=black
                                    [
                                        COLOR \citep{yoo2023advancing}
                                        , leaf, text width=12.7em, color=cyan!100, fill=cyan!15, very thick, text=black
                                    ]
                                ]
                                [
                                    Secret key per message, color=cyan!100, fill=cyan!15, very thick, text=black
                                    [
                                        \citep{fernandez2023three}
                                        , leaf, text width=12.7em, color=cyan!100, fill=cyan!15, very thick, text=black
                                    ]
                                ]
                            ]
                        ]
                    ]
                ]
            ]
        \end{forest}
    }
    \caption{Taxonomy on detection of LLMs-generated content. We list the most representative approaches for each kind of method.}
    \label{tab:categorization_of_reasoning}
\end{figure*}
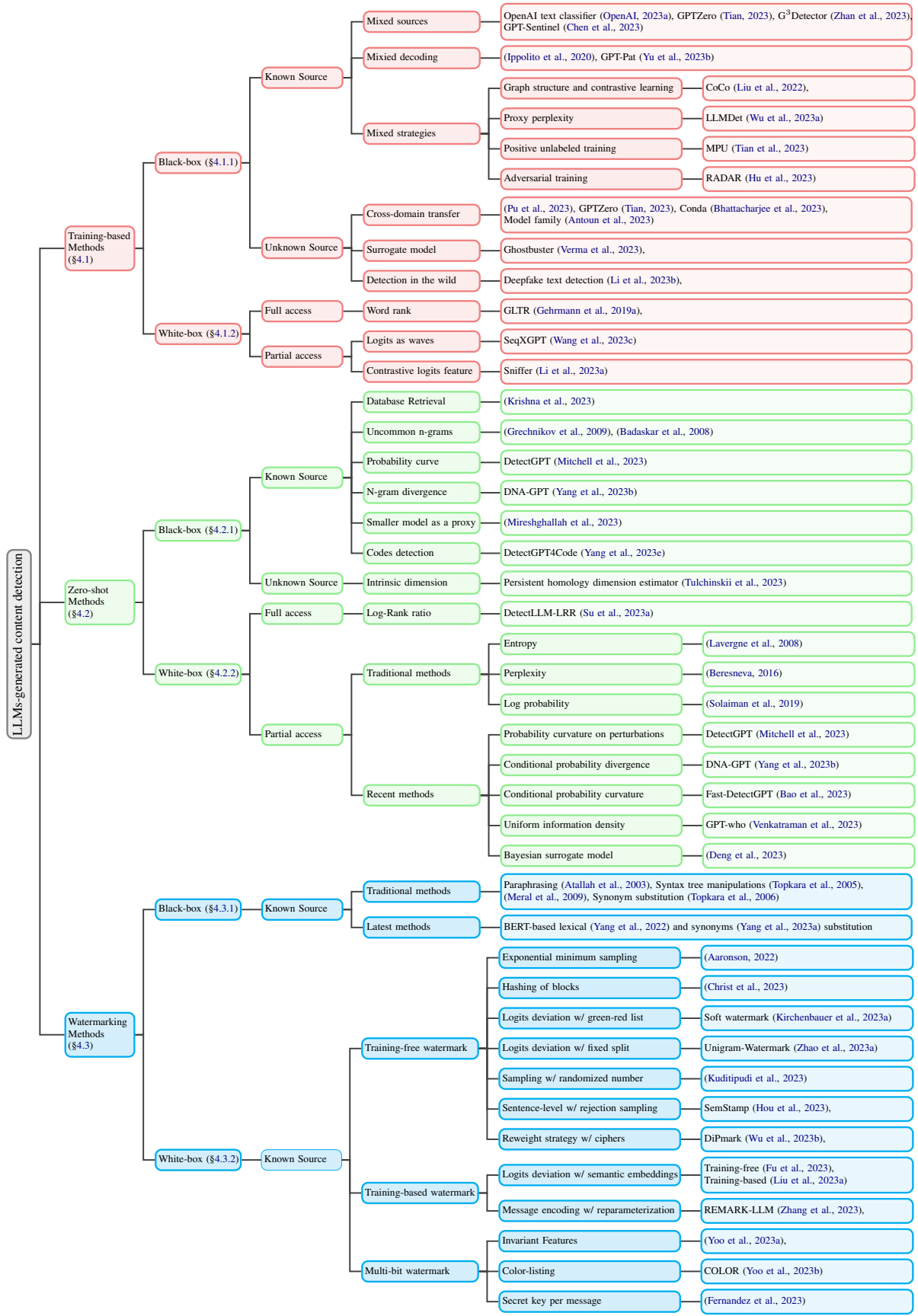
\section{Detection Methodologies}\label{sec:Methodologies}
In this section, we delve into further details about the detection algorithms. Based on their distinguishing characteristics, existing detection methods can be categorized into three classes: 1) Training-based classifiers, which typically fine-tune a pre-trained language model on collected binary data - both human and AI-generated text distributions. 2) Zero-shot detectors leverage the intrinsic properties of typical LLMs, such as probability curves or representation spaces, to perform self-detection. 3) Watermarking involves hiding identifying information within the generated text that can later be used to determine if the text came from a specific language model, rather than detecting AI-generated text in general. We summarize the representative approaches in Figure \ref{tab:categorization_of_reasoning} as classified by the scenarios listed in Section \ref{sec: Scenarios}.
\vspace{-0.1cm}
\subsection{ Training-based \includegraphics[scale=0.1]{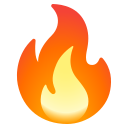} }\label{sec:training-method}
\vspace{-0.1cm}
The earlier work of training a detection classifier focuses on fake review \citep{bhagat2013paraphrase}, fake news \citep{zellers2019defending} or small models \citep{solaiman2019release, bakhtin2019real, uchendu-etal-2020-authorship} detection. Subsequently, growing interest in this line of research turns to detecting high-quality text brought by LLMs.

\subsubsection{Black-box}\label{sec:training-black-box}
The first line of work focuses on black-box detection. \textit{When the model source is known}, some work use the text generated by \textbf{ \tikz[baseline=(char.base)]{\node[shape=circle,draw,inner sep=1pt] (char) {1};} mixed sources} and subsequently train a classifier together for detection. For example,
OpenAI \citep{AITextClassifier} collects text generated from different model families and trains a robust detector for detection text with more than 1,000 tokens. GPTZero \citep{GPTZero} also collects their human-written text spans student-written articles, news articles, and Q\&A datasets spanning multiple disciplines from a variety of LLMs. Similarly, G$^3$Detector \citep{zhan2023g3detector} claims to be a general GPT-Generated text detector by finetuning RoBERTa-large \citep{liu2019roberta} and explores the effect of the use of synthetic data on the training process. GPT-Sentinel \citep{chen2023gpt} trains the RoBERTa and T5 \citep{raffel2020exploring} classifiers on their constructed dataset OpenGPTText. \textbf{ \tikz[baseline=(char.base)]{\node[shape=circle,draw,inner sep=1pt] (char) {2};} Mixed decoding} is also utilized by incorporating text generated with different decoding parameters to account for the variance. \citet{ippolito-etal-2020-automatic} find that, in general, discriminators transfer poorly between decoding strategies, but training on a mix of data can help. GPT-Pat \citep{yu2023gpt} train a siamese network to compute the similarity between the original text and the re-decoded text. Besides, \textbf{ \tikz[baseline=(char.base)]{\node[shape=circle,draw,inner sep=1pt] (char) {3};} mixed strategies} involves additional information, such as graph structure and contrastive learning in CoCo \citep{liu2022coco}, proxy model perplexity in LLMDet \citep{wu2023llmdet}, positive unlabeled training in MPU \citep{tian2023multiscale} and adversarial training in RADAR \citep{hu2023radar}.

On the other hand, \textit{when the source model is unknown}, OpenAI text classifier \citep{AITextClassifier} and GPTZero \citep{GPTZero} still works by \textbf{ \tikz[baseline=(char.base)]{\node[shape=circle,draw,inner sep=1pt] (char) {1};} cross-domain transfer}. Other works like \citep{pu2023zero, Antoun2023FromTT}, Conda \citep{Bhattacharjee2023ConDACD} also rely on the zero-shot
generalization ability of detectors trained on a variety of model families and tested on unseen models. 
Besides, Ghostbuster \citep{verma2023ghostbuster} directly uses outputs from known \textbf{ \tikz[baseline=(char.base)]{\node[shape=circle,draw,inner sep=1pt] (char) {2};} surrogate model} as the signal for training a classifier to detect unknown model. 
Additionally, \textbf{ \tikz[baseline=(char.base)]{\node[shape=circle,draw,inner sep=1pt] (char) {3};} detection in the wild} \citep{li2023deepfake} contributes a wild testbed by gathering texts from various human writings and deepfake texts generated by different LLMs for detection without knowing their sources. 

\subsubsection{White-box}\label{sec:training-white-box}
The second kind of work lies in the white-box situation when the model's full or partial parameters are accessible. For example, when we have full access to the model, GLTR \citep{gehrmann2019gltr} trains a logistic regression over absolute word ranks in each decoding step. When only partial information like the model output logits are available, SeqXGPT \citep{wang2023seqxgpt} introduce a sentence-level detection challenge by synthesizing a dataset that contains documents that are polished with LLMs and propose to detect it with logits as waves from white-box LLMs. Sniffer \citep{li2023origin} utilizes the contrastive logits between models as a typical feature for training to perform both detection and origin tracking.

%%%%%%%%%%%%%%%%%% Zero-Shot %%%%%%%%%%%%%%%%%%
\subsection{Zero-Shot \includegraphics[scale=0.1]{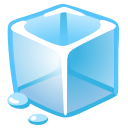} }\label{sec:Zero-Shot}
In the zero-shot setting, we do not require extensive training data to train a discriminator. Instead, we can leverage the inherent distinctions between machine-generated and human-written text, making the detector training-free. The key advantage of training-free detection is its adaptability to new data distributions without the need for additional data collection and model tuning. It's worth noting that while watermarking methods can also be considered zero-shot, we treat them as an independent track.
Previous work utilizes entropy \citep{lavergne2008detecting}, average log-probability score \citep{solaiman2019release}, perplexity \citep{beresneva2016computer}, uncommon n-gram
frequencies \citep{grechnikov2009detection, badaskar2008identifying} obtained from a language model as the judge for determining its origin. However, those simple features fail as LLMs are becoming diverse and high-quality text generators. Similarly, there are also black- and white-box detection, as summarized below.

\subsubsection{Black-Box}\label{zero-shot: black}
\textit{When the source of the black-box model is known}, DNA-GPT \citep{yang2023dna} achieves superior performance by utilizing N-Gram divergence between the continuation distribution of re-prompted text and the original text.
Besides, DetectGPT \citep{mitchell2023detectgpt} also investigates using another surrogate model to replace the source model but achieves unsatisfactory results. In contrast, \citet{mireshghallah2023smaller} proves that a smaller surrogate model like OPT-125M \citep{zhang2022opt} can serve as a universal black-box text detector, achieving close or even better detection performance than using the source model. Additionally, \citet{krishna2023paraphrasing} suggests building a database of generated text and detecting the target text by comparing its semantic similarity with all the text stored in the database. Finally, DetectGPT4Code \citep{yang2023zero} also investigates detecting codes generated by ChatGPT through a proxy small code generation models by conditional probability divergence and achieves significant improvements on code detection tasks.

\textit{When the source of the model is unknown}, PHD \citep{tulchinskii2023intrinsic} observes that real text exhibits a statistically higher intrinsic dimensionality compared to machine-generated texts across various reliable generators by employing the Persistent Homology Dimension Estimator (PHD) as a means to measure this intrinsic dimensionality, combined with an additional encoder like Roberta to facilitate the estimation process.

\subsubsection{White-Box}\label{zero-shot: white}
\textit{When the partial access to the model is given}, traditional methods use the features such as entropy \citep{lavergne2008detecting}, average log-probability score \citep{solaiman2019release} for detection. However, these approaches struggle to detect text from the most recent LLMs. Then, the pioneer work DetectGPT \citep{mitchell2023detectgpt} observes that LLM-generated text tends to occupy negative curvature regions of the model's log probability function and leverages the curvature-based criterion based on random perturbations of the passage. DNA-GPT \citep{yang2023dna} utilizes the probability difference between the continuous distribution among re-prompted text and original text and achieves state-of-the-art performance. Later, \citet{deng2023efficient} improves the efficiency of DetectGPT with a Bayesian surrogate model by selecting typical samples based on Bayesian uncertainty and interpolating scores from typical samples to other ones. Furthermore, similar to DNA-GPT \citep{yang2023dna} on using the conditional probability for discrimination, Fast-DetectGPT \citep{bao2023fast} builds an efficient zero-shot detector by replacing the probability in DetectGPT with conditional probability curvature and witnesses significant efficiency improvements. Additionally, GPT-who \citep{venkatraman2023gptwho} utilizes Uniform Information Density (UID) based features to model the unique statistical signature of each LLM and human author for accurate authorship attribution.

\textit{When the full access to the model is given}, \citet{su2023detectllm} leverages the log-rank information for zero-shot detection through one fast and efficient DetectLLM-LRR (Log-\textbf{L}ikelihood \textbf{L}og-Rank \textbf{r}atio) method, and another more accurate DetectLLM-NPR (\textbf{N}ormalized \textbf{p}erturbed log \textbf{r}ank) method, although slower due to the need for perturbations.

%%%%%%%%%%%%%%%%%% Watermarking %%%%%%%%%%%%%%%%%%
\subsection{Watermarking \includegraphics[scale=0.1]{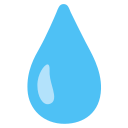} }\label{sec:Watermarking}
Text watermarking injects algorithmically detectable patterns into the generated text while ideally preserving the quality and diversity of language model outputs. Although the concept of watermarking is well-established in vision, its application to digital text poses unique challenges due to the text's discrete and semantic-sensitive nature \citep{Kutter2000InformationHT}. 
Early works are edit-based methods that modify a pre-existing text. The earliest work can be dated back to \citet{atallah2001natural}, which designs a scheme for watermarking natural language text by embedding small portions of the watermark bit string in the syntactic structure of the text, followed by paraphrasing \citep{atallah2003natural}, syntax tree manipulations \citep{topkara2005natural, meral2009natural} and synonym substitution \citep{topkara2006hiding}. 
Besides, text watermarking has also been used for steganography and secret communication \citep{fang-etal-2017-generating, ziegler-etal-2019-neural, abdelnabi21oakland}, and intellectual property protection \citep{he2022protecting, he2022cater, zhao-etal-2022-distillation, Zhao2023ProtectingLG}, but this is out the scope of this work. In light of growing ethical considerations, text watermarking has been increasingly used to ascertain the origin of textual content and detect AI-generated content \citep{grinbaum2022ethical}. The primary focus of this paper is on the use of text watermarking to detect AI-generated text.

In general, watermarking for text detection can also be classified into white-box and black-box watermarking. Watermarking is designed to determine whether the text is coming from a specific language model rather than universally detecting text generated by any potential model. As such, knowledge of the model source is always required in text watermarking for detection.

\subsubsection{Black-Box Watermarking}\label{watermark:black} 
In black-box setting, such as API-based applications, the proprietary nature of the language models used by LLM providers precludes downstream users from accessing the sampling process for commercial reasons. Alternatively, a user may wish to watermark human-authored text via post-processing. In such cases, black-box watermarking aims to automatically manipulate generated text to embed watermarks readable by third parties. Traditional works designed complex linguistic rules such as paraphrasing \citep{atallah2003natural}, syntax tree manipulations \citep{topkara2005natural, meral2009natural} and synonym substitution \citep{topkara2006hiding}, but lack scalability. 
Later work turns to pre-trained language models for efficient watermarking. For example, \citet{yang2022tracing} proposes a natural language watermarking scheme based on context-aware lexical substitution (LS). Specifically, they employ BERT \citep{devlin-etal-2019-bert} to suggest LS candidates by inferring the semantic relatedness between the candidates and the original sentence.
\citet{yang2023watermarking} first defines a binary encoding function to compute a random binary encoding corresponding to a word. The encodings computed for non-watermarked text conform to a Bernoulli distribution, wherein the probability of a word representing bit-1 is approximately 0.5. To inject a watermark, they alter the distribution by selectively replacing words representing bit-0 with context-based synonyms that represent bit-1. A statistical test is then used to identify the watermark.

\subsubsection{White-Box Watermarking}\label{watermark:white}
The most popular \textbf{ \tikz[baseline=(char.base)]{\node[shape=circle,draw,inner sep=1pt] (char) {1};} training-free} watermark directly manipulates the decoding process when the model is deployed. In the efforts of watermarking GPT outputs, \citet{aaronson2022my} works with OpenAI to first develop a technique for watermarking language models using exponential minimum sampling to sample text from the model, where the inputs to the sampling mechanism are a hash of the previous $k$ consecutive tokens through a pseudo-random number generator. By Gumbel Softmax \citep{jang2016categorical} rule, their method is proven to ensure guaranteed quality. Besides, \citet{christ2023undetectable} provides the formal definition and construction of undetectable watermarks. Their cryptographically inspired watermark design proposes watermarking blocks of text from a language model by hashing each block to seed a sampler for the next block. However, there are only theoretical concepts for this method without experimental results.
Another pioneering work of training-free watermark \citep{kirchenbauer2023watermark} embeds invisible watermarks in the decoding process by dividing the vocabulary into a ``green list'' and a ``red list'' based on the hash of prefix token and subtly increases the probability of choosing from the green list. Then, a third party, equipped with knowledge of the hash function and random number generator, can reproduce the green list for each token and monitor the violation of the green list rule. Subsequently, \citet{Zhao2023ProvableRW} simplifies the scheme by consistently using a fixed green-red list split, showing that the new watermark persists in guaranteed generation quality and is more robust against text editing. \citet{Kuditipudi2023RobustDW} create watermarks that are distortion-free by utilizing randomized watermark keys to sample from token probability distribution by inverse transform sampling and exponential minimum sampling. \citet{hou2023semstamp} propose a sentence-level semantic watermark based on locality-sensitive hashing (LSH), which partitions the semantic space of sentences. The advantage of this design is its enhanced robustness against paraphrasing attacks. DiPmark \citep{wu2023dipmark} is an unbiased distribution-preserving watermark that preserves the original token distribution during watermarking and is robust to moderate changes of tokens by incorporating a novel reweight strategy, combined with a hash
function that assigns unique i.i.d. ciphers based on the context.
Drawn on the drawbacks of random green-red list splitting, \citet{fu2023watermarking} uses input sequence to get semantically related tokens for watermarking to improve certain conditional generation tasks.

Despite training-free watermarking, text watermarks can also be injected through pre-inference training or post-inference training: \textbf{ \tikz[baseline=(char.base)]{\node[shape=circle,draw,inner sep=1pt] (char) {2};} training-based watermark}.
One example of pre-inference training is REMARK-LLM \citep{zhang2023remark}, which injects the watermark by a message encoding module to generate a dense token distribution, following a message decoding module to extract messages from the watermarked textual and reparameterization is used as a bridge to connect the dense distribution with tokens’ one-hot encoding. The drawback is that training is required on source data and might not generalize well to unseen text data.
On the contrary, post-inference training involves adding a trained module to assist in injecting watermarks during inference. For instance, \citet{liu2023semantic} proposes a semantic invariant robust watermark for LLMs, by utilizing another embedding LLM to generate semantic embeddings for all preceding tokens. However, it is not training-free since these semantic embeddings are transformed into the watermark logits through their trained watermark model.

Despite from 0-bit watermark, there is also \textbf{ \tikz[baseline=(char.base)]{\node[shape=circle,draw,inner sep=1pt] (char) {3};} multi-bit watermarking}. For example, \citet{yoo-etal-2023-robust} designs a multi-bit watermarking following a well-known proposition from image watermarking that identifies natural language features invariant to minor corruption and proposes a corruption-resistant infill model. COLOR \citep{yoo2023advancing} subsequently designs another multi-bit watermark by embedding traceable multi-bit information during language model generation while allowing zero-bit detection simultaneously.
\citet{fernandez2023three} also consolidates watermarks for LLMs through more robust statistical tests and multi-bit watermarking.

\subsection{Commercial Tool}
Despite from academic research, AI text detection also draws considerable attention from commercial companies.
Table \ref{tab: tool} summarizes the popular commercial detectors. Although the majority of them simultaneously claim to be the most accurate AI detectors on the homepage of their website, it is essential to evaluate their performance based on various factors such as accuracy, speed, robustness, and compatibility with different platforms and frameworks. Regrettably, a dearth of articles exists that explicitly delve into the comparative analysis of the aforementioned properties among popular commercial detectors. 

\begin{table*}[htbp]
  \centering
\resizebox{1.\textwidth}{!}
{
  \begin{tabular}{c|c|c|c}
    \hline
    \textbf{Product Name} & \textbf{Website} & \textbf{Price} & \textbf{API available}\\
    \hline
    Originality.AI & \url{https://app.originality.ai/api-access} & \$0.01/100 words & Yes \\ 
    \hline
    Quil.org & \url{https://aiwritingcheck.org/} & Free website version & No \\
    \hline
    Sapling & \url{https://sapling.ai/ai-content-detector} & 1 million chars at \$25/month & Yes  \\
    \hline
    OpenAI text classifier & \url{https://openai-openai-detector.hf.space/} & Free website version & Yes  \\
    \hline
    Crossplag & \url{https://crossplag.com/ai-content-detector/} & Free website version & No \\
    \hline
    GPTZero & \url{https://gptzero.me/} & 0.5 million words at \$14.99/mo & Yes  \\
    \hline
    ZeroGPT & \url{https://www.zerogpt.com/} & Free website version & No  \\
    \hline
    CopyLeaks & \url{https://copyleaks.com/ai-content-detector} & 25000 words at \$10.99/Month & No \\ % , support batch upload
    \hline
    
  \end{tabular}
  }
  \caption{A summary of popular commercial tools to detect AI-generated text. }\label{tab: tool}
\end{table*}

\vspace{-0.1cm}
\section{Detection Attack}\label{sec: attack}
\vspace{-0.1cm}
Despite the progress of detection work, there are also continuous efforts to evade existing detectors, and we summarize the main streams in this section.
\vspace{-0.1cm}
\subsection{Paraphrasing Attack }
\vspace{-0.1cm}
Paraphrasing could be performed by human writers or other LLMs, and even by the same source model. Paraphrasing can also undergo several rounds, influenced by a mixture of different models. Current research mostly focuses on the simple paraphrase case where another model rewrites a machine-generated text for one round. For instance, \citet{krishna2023paraphrasing} trains a T5-11b model for paraphrasing text and discovers that all detectors experience a significant drop in quality when faced with paraphrased text. Additionally, simple paraphrasing attacks involve word substitutions \citep{shi2023red}. Moreover, paraphrasing can also be achieved through translation attacks. However, conducting more in-depth analysis and research on complex paraphrasing techniques in the future is crucial. \citet{becker2023paraphrase} systemically examines different classifiers encompassing both classical approaches and Transformer techniques for detecting machine (like T5) or human paraphrased text.

\subsection{Adversarial Attack}
Though the adversarial attack is popular for general NLP tasks \citep{alzantot2018generating}, there has been little work specifically addressing adversarial attacks on detectors for LLM-generated content. However, we can consider the following two types of attacks for further investigation and exploration:

\textit{Adversarial Examples:} Attackers can generate specially crafted inputs by making subtle modifications to the text that fool the AI text detectors while remaining mostly unchanged to human readers \citep{shi2023red}. These modifications can include adding or removing certain words or characters, introducing synonyms, or leveraging linguistic tricks to deceive the detector. Evasion attacks aim to manipulate the AI text detector's behavior by exploiting its vulnerabilities. Attackers can use techniques such as obfuscation, word permutation, or introducing irrelevant or misleading content to evade detection. The goal is to trigger false negatives and avoid being flagged as malicious or inappropriate.

\textit{Model Inversion Attacks:} Attackers can launch model inversion attacks by exploiting the responses of AI text detectors. They might submit carefully crafted queries and observe the model's responses to gain insights into its internal workings, architecture, or training data, which can be used to create more effective attacks or subvert the system's defenses.

\subsection{Prompt Attack}
Current LLMs are vulnerable to prompts \citep{zhu2023promptbench}, thus, users can utilize smartly designed prompts to evade established detectors. 
For example, \citet{shi2023red} examines instructional prompt attacks by perturbing the input prompt to encourage LLMs to generate texts that are difficult to detect.
\citet{lu2023large} also show that LLMs can be guided to evade AI-generated text detection by a novel substitution-based In-Context example Optimization method (SICO) to automatically generate carefully crafted prompts, enabling ChatGPT to evade six existing detectors by a significant 0.54 AUC drop on average. 
Nevertheless, limited attention has been devoted to this topic, indicating a notable research gap that merits significant scholarly exploration in the immediate future.
Notably, a recent work \citep{chakraborty2023counter} introduces the Counter Turing Test (CT2), a benchmark consisting of techniques aiming to evaluate the robustness of existing six detection techniques comprehensively. Their empirical findings unequivocally highlight the fragility of almost all the proposed detection methods under scrutiny. 
Despite the hard prompt attack, \citet{kumarage2023reliable} first creates an evasive soft prompt tailored to a specific PLM through prompt tuning; and then, they leverage the transferability of soft prompts to transfer the learned evasive soft prompt from one PLM to another and find the universal efficacy of the evasion attack.

\vspace{-0.1cm}
\section{Challenges}\label{sec: challenge}
\vspace{-0.1cm}
\subsection{Theorical Analysis}
Inspired by the binary hypothesis test in \citep{polyanskiy2022information}, \citep{sadasivan2023can} claims that machine-generated text will become indistinguishable as the total variance between the distributions of human and machine approaches zero. In contrast, \citet{chakraborty2023possibilities} demonstrates that it is always possible to distinguish them by curating more data to make the detection of AUROC increase exponentially with the number of training instances. Additionally, DNA-GPT \citep{yang2023dna} demonstrates the difficulty of obtaining a high TPR while maintaining a low FPR.
Nevertheless, a dearth of theoretical examination persists regarding the disparities in intrinsic characteristics between human-written language and LLMs. Scholars could leverage the working mechanisms of GPT models to establish a robust theoretical analysis, shedding light on detectability and fostering the development of additional detection algorithms.

\vspace{-0.1cm}
\subsection{LLM-Generated Code Detection}
\vspace{-0.1cm}
Previous detectors usually only focus on the text, but LLMs-generated codes also show increasing quality (see a recent survey \citep{zan2022neural}). Among the first, \citet{lee2023wrote} found that previous watermarking \citep{kirchenbauer2023watermark} for text does not work well in terms of both detectability and generated code quality. It is evidenced that low entropy persists in generated code \citep{lee2023wrote}, thus, the decoding process is more deterministic. They thus adapt the text watermarks to code generation by only injecting watermarks to tokens with higher entropy than a given threshold and achieve more satisfactory results. Code detection is generally believed to be even harder than text detection due to its shorter length, low entropy, and non-natural language properties. DetectGPT4Code \citep{yang2023zero} detects codes generated by ChatGPT by using a proxy code model to approximate the logits on the conditional probability curve and achieves the best results over previous detectors.

\vspace{-0.2cm}
\subsection{Model Sourcing}
\vspace{-0.1cm}
Model sourcing \citep{yang2023dna}, is also known as origin tracking \citep{li2023origin} or authorship attribution \citep{uchendu-etal-2020-authorship}.
Unlike the traditional distinction between human and machine-generated texts, it focuses on identifying the specific source model from a pool of models, treating humans as a distinct model category. With the fast advancement of LLMs from different organizations, it is vital to tell which model or organization potentially generates a certain text. This has practical applications, particularly for copyright protection.
Consequently, we believe that in the future, it may become the responsibility of organizations releasing powerful LLMs to determine whether a given text is a product of their system. Previous work either \citep{li2023origin} trains a classifier or utilizes the intrinsic genetic properties \citep{yang2023dna} to perform model sourcing, but still can not handle more complicated scenarios. GPT-who \citep{venkatraman2023gptwho} utilizes Uniform Information Density (UID) based features to model the unique statistical signature of each LLM and human author for accurate authorship attribution.

\vspace{-0.1cm}
\subsection{Bias}
It has been found that current detectors tend to be biased against non-native speakers \citep{liang2023gpt}. Also, \citet{yang2023dna} found that previous detection tools often perform poorly on other languages other than English. Besides, current research usually focuses on the detection of text within a certain length, thus showing bias against the shorter text. How to ensure the integrity of detectors under various scenarios without showing bias against certain groups is of central importance.

\vspace{-0.1cm}
\subsection{Generalization}
Currently, the most advanced LLMs, like ChatGPT, are getting actively updated, and OpenAI will make a large update every three months. 
How to effectively adapt existing detectors to the updated LLMs is of great importance. For example, \citet{tu2023chatlog} records the ChatLog of ChatGPT's response to long-form generation every day in one month, observes performance degradation of the Roberta-based detector, and also finds some stable features to improve the robustness of detection. As LLMs continuously benefit from interacting with different datasets and human feedback, exploring ways to effectively and efficiently detect their generations remains an ongoing research area.
Additionally, \citet{kirchenbauer2023reliability} investigates the reliability of watermarks for large language models and claims that watermarking is a reliable solution under human paraphrasing and various attacks at the context length of around 1000. 
\citet{pu2023zero} examines the zero-shot generalization of machine-generated text detectors and finds that none of the detectors can generalize to all generators.
All those findings reveal the difficulty of reliable generalization to unseen models or data sources of detection.

\vspace{-0.1cm}
\section{Future Outlook}\label{sec: outlook}
\vspace{-0.1cm}
The detection of LLM-generated content is an evolving field. Here, we list some potential avenues for future work (details are included in Appendix \ref{app:future}): 1). robust and scalable detection techniques; 2). rigorous and standard evaluation; 3). fine-grained detection; 4). user education and awareness; 5). transparency and explainability.

\vspace{-0.1cm}
\section{Conclusion}\label{sec: conclusion}
\vspace{-0.1cm}
We comprehensively survey LLMs-generated content detection over existing task formulation, benchmark datasets, evaluation metrics, and different detection methods under various scenarios. 
We also point out existing challenges, such as diversified attack approaches, and share our views on the future of detection. 
We hope our survey can help the research community quickly learn the progress of detection methodologies and challenges and potentially inspire more ideas in the urgent need for reliable detectors. 

\section*{Limitations}
Despite conducting a comprehensive literature review on AI-generated content detection, we acknowledge the potential for omissions due to incomplete searches.

\section*{Ethics Statement}
The utilization of AI detection presents significant ethical considerations, particularly when it comes to the detection of plagiarism among students. Misclassifications in this context can give rise to substantial concerns. This survey aims to summarize the current techniques employed in this field comprehensively. However, it is important to note that no flawless detectors have been developed thus far. Consequently, users should exercise caution when interpreting the detection outcomes, and it should be understood that we cannot be held accountable for any inaccuracies or errors that may arise.

% \clearpage
% \newpage

\bibliography{anthology,custom}
\bibliographystyle{acl_natbib}

\appendix
\newpage
\section{Appendix}

\subsection{News Reports}\label{app: news}
We summarize several influential news on the false use of AI detectors and concerns brought by AI-generated information.\\
1. International students are concerned their original writing is being flagged as AI-generated text. \href{https://t.co/c4fldWEf6f}{link} \\
2. Professor Flunks All His Students After ChatGPT Falsely Claims It Wrote Their Papers. \href{https://www.rollingstone.com/culture/culture-features/texas-am-chatgpt-ai-professor-flunks-students-false-claims-1234736601/}{link} \\
3. China reports first arrest over fake news generated by ChatGPT. \href{https://www.reuters.com/technology/china-reports-first-arrest-over-fake-news-generated-by-chatgpt-2023-05-10/}{link} \\
4. Professors have a summer assignment: Prevent ChatGPT chaos in the fall. \href{https://www.washingtonpost.com/technology/2023/08/13/ai-chatgpt-chatbots-college-cheating/}{link} \\
5. AI makes plagiarism harder to detect, argue academics – in paper written by chatbot. \href{https://www.theguardian.com/technology/2023/mar/19/ai-makes-plagiarism-harder-to-detect-argue-academics-in-paper-written-by-chatbot}{link} \\
6. How AI Could Take Over Elections—And Undermine Democracy. \href{https://www.scientificamerican.com/article/how-ai-could-take-over-elections-and-undermine-democracy/}{link} \\

\subsection{Related Survey}
In the literature, there are some other surveys on this topic. For example, \citet{jawahar-etal-2020-automatic} discusses the detection of small language models. \citet{tang2023science} provides an overview of previous detection methods but does not fully cover the recent progress in the era of LLMs. Very recently, \citet{crothers2022machine} surveys threat models and detection methods but also summarizes previous detection methods rather than the latest progress with LLMs. Unlike them, our work aims to fill this gap by providing the first comprehensive survey about detection, attack, and benchmarks, especially focusing on detecting LLMs like ChatGPT. Thus, our survey includes the most advanced approaches. 

\citet{dhaini2023detecting} gives a survey of the state of detecting only ChatGPT-Generated text but ignores various detection methods on other models.

\subsection{Datasets}
\label{sec:datasets}
\begin{itemize}[leftmargin=*]
\setlength{\itemsep}{0.2pt}
\setlength{\parsep}{0.2pt}
\setlength{\parskip}{0.2pt}
\item \citet{uchendu-etal-2021-turingbench-benchmark} presents the TURINGBENCH benchmark for Turing Test and Authorship Attribution across 19 language models.
\item HC3 \citep{guo2023close} collectes the Human ChatGPT Comparison Corpus (HC3) with both long- and short-level documents from ELI5 \citep{fan-etal-2019-eli5}, WikiQA \citep{yang-etal-2015-wikiqa}, Crawled Wikipedia, Medical Dialog \citep{chen2020meddialog}, and FiQA \citep{maia201818}.

\item CHEAT \citep{yu2023cheat} provides 35,304 synthetic academic abstracts, with Generation, Polish, and Mix as prominent representatives.

\item Ghostbuster \citep{verma2023ghostbuster} provides a detection benchmark that covers student essays, creative fiction, and news at document-level detection and paragraph-level.

\item OpenGPTText \citep{chen2023gpt} consists of 29,395 rephrased content generated using ChatGPT, originating from OpenWebText \citep{Gokaslan2019OpenWeb}.

\item M4 \citep {wang2023m4} is a large-scale benchmark covering multi-generator, multi-domain, and multi-lingual corpus for machine-generated text detection.

\item MULTITuDE \citep{macko2023multitude} Large-Scale Multilingual Machine-Generated Text Detection Benchmark comprising 74,081 authentic and machine-generated texts in 11 languages generated by 8 multilingual LLMs. They find that the most currently available black-box methods do not work in multilingual settings.

\item MGTBench \citep{he2023mgtbench} focuses on ChatGPT-generated content on: TruthfulQA \citep{lin-etal-2022-truthfulqa}, SQuAD \citep{rajpurkar-etal-2016-squad} and NarrativeQA \citep{kocisky-etal-2018-narrativeqa}.

\item SAID(Social media AI Detection) \citet{Cui2023WhoST} is curated for real AI-generate text from popular social media platforms like Zhihu and Quora, and conducting detection tasks on actual social media platforms prove to be more
challenging compared to traditional simulated AI-text detection.

\item HC3 Plus \citep{su2023hc3} is a more extensive and comprehensive dataset that considers more types of tasks, considering tasks such as summarization, translation, and paraphrasing to possess semantic-invariant properties and are more difficult to detect. 
\end{itemize}

\subsection{Future Outlooks}\label{app:future}
Details on the future outlook are as follows.
\begin{itemize}[leftmargin=*]
\setlength{\itemsep}{0.2pt}
\setlength{\parsep}{0.2pt}
\setlength{\parskip}{0.2pt}

\item \textit{Robust and Scalable Detection Techniques}: Current LLMs are getting constant improvements from big tech companies. Thus, the development of advanced algorithms and detection techniques capable of accurately identifying LLM-generated content in real time is a priority. Future research should focus on improving the accuracy, robustness to attacks, and scalability of detection methods to keep up with the increasing volume and complexity of LLM-generated content.

\item \textit{Rigorous and Standard Evaluation}: As discussed in Section \ref{02: datasets}, current evaluation faces data contamination issues; either the LLMs or the detectors might encounter the human data in their training stage.
Besides, the evaluation benchmark also varies. The detection results affect the length, prompting methods, and adopted datasets. However, unlike traditional machine learning tasks where one benchmark can be used for a long period, how to avoid any potential data contamination is very critical. 

\item \textit{Fine-grained Detection}: LLM-generated content can vary in its intentions, ranging from malicious propaganda to unintentional misinformation. 
Future work should explore approaches that can detect and differentiate between various categories of LLM-generated content, allowing for more tailored interventions and countermeasures.

\item \textit{User Education and Awareness}: Educating users about the existence and capabilities of LLMs detectors is essential. For example, in Appendix \ref{app: news}, we show some reported misuse of AI detectors in education.
Future work should focus on raising awareness among users about the reliability of detection tools. This can empower users to make more informed decisions and mitigate the impact of deceptive or misleading decisions.

\item \textit{AI Regulations}: As LLMs become more sophisticated, the ethical implications of their usage in generating deceptive content become increasingly important. Future research should explore ethical frameworks and guidelines for the responsible development and deployment of LLMs while considering the potential consequences and risks associated with their misuse. 

\item \textit{Transparency and Explainability}: Enhancing the transparency and explainability of LLM-generated content detection algorithms is crucial for building trust and understanding among users. For example, \citet{yang2023dna} uses the non-trivial N-gram overlaps to support the detection results. But currently, most detectors can only give a predictive probability, with no clues about evidence.
Future work should focus on developing techniques that can provide explanations or evidence for the classification decisions made by detection systems, enabling users to understand the rationale behind content identification better.
\end{itemize}\label{sec:appendix}

\end{document}